%% file: output.tex
\documentclass[10pt]{article} 
\usepackage[preprint]{tmlr}

\input{math_commands.tex}

\usepackage{hyperref}
\usepackage{url}

\usepackage{graphicx}
\usepackage{cleveref}
\usepackage{listings}
\usepackage{xcolor}
\usepackage{nicefrac}
\usepackage{enumitem}  

\usepackage{lstfiracode}

\usepackage[frozencache,cachedir=.]{minted}

\usepackage[most,minted]{tcolorbox}

\usepackage{float}
\restylefloat{table}

\title{SequenceLayers: Sequence Processing and Streaming \\Neural Networks Made Easy}

\author{\name RJ Skerry-Ryan, Julian Salazar, Soroosh Mariooryad, David Kao, \email rjryan@google.com \\
\name Daisy Stanton, Eric Battenberg, Matt Shannon, Ron Weiss,\\
\name Robin Scheibler, Jonas Rothfuss, Tom Bagby \\
      \addr Google DeepMind}

\def\month{MM}  
\def\year{YYYY} 
\def\openreview{\url{https://openreview.net/forum?id=XXXX}} 

\begin{document}

\maketitle

\begin{abstract}
We introduce a neural network layer API and library for sequence modeling, designed for easy creation of sequence models that can be executed both layer-by-layer (e.g., teacher-forced training) and step-by-step (e.g., autoregressive sampling). 
To achieve this, layers define an explicit representation of their \texttt{state} over time (e.g., a Transformer KV cache, a convolution buffer, an RNN hidden state), and a \texttt{step} method that evolves that state, tested to give identical results to a stateless layer-wise invocation. 
This and other aspects of the SequenceLayers contract enables complex models to be immediately streamable, mitigates a wide range of common bugs arising in both streaming and parallel sequence processing, and can be implemented in any deep learning library.
A composable and declarative API, along with a comprehensive suite of layers and combinators, streamlines the construction of production-scale models from simple streamable components while preserving strong correctness guarantees.
Our current implementations of SequenceLayers (JAX, TensorFlow 2) are available at \url{https://github.com/google/sequence-layers}.
\end{abstract}

\section{Introduction}

Neural network models that perform sequential processing have become central to deep learning, from sequence-to-sequence models for machine translation, text-to-speech, and speech recognition, to real-time variants which perform live translation, dialogue generation, and transcription, to next-token generative models for language (LLMs), media, and beyond. However, sequence processing has some common pitfalls:
\begin{itemize}
\item \textbf{Batching sequences of unequal length.} One must track and mask invalid timesteps, and verify that all layers are invariant to padding, including pooling, downsampling, and upsampling operations.
\item \textbf{Causality constraints.} Modern architectures like the Transformer \citep{vaswani2017transformer} often have an efficient parallel ("teacher-forced") training code path implemented separately from the autoregressive sampling path used during inference. 
Both code paths must be implemented in a way that avoids causality violations.
\item \textbf{Offline vs.\ streaming inference mismatch.} Offline (parallel) inference is often implemented via masking on e.g.\ attention weights. Converting this to a streaming setting where masking is implicit can require re-implementation due to considerations like lookahead windows and memory constraints, causing training and inference disparities.
\item \textbf{Unnecessary coupling of architecture and algorithm.} Lack of abstraction of architectural details typically leads to unnecessary coupling. For example an "\texttt{AutoregressiveTransformer}" couples both the model architecture (Transformer) and the algorithm (factoring the joint probability of sequences autoregressively via the chain rule). The details of autoregressive modeling (mapping from previous timepoints to a parameterized distribution, and sampling from that distribution) are independent of the choice of architecture (e.g., Transformer, RNN, state-space model, ...), yet are often unnecessarily coupled which can hinder experimentation and produce research technical debt.
\end{itemize}

To improve the experience of developing sequence models, we propose SequenceLayers, a lightweight \textit{layer} library for sequential neural networks. It has \textbf{three core features}:
\begin{enumerate}
    \item \textbf{Streamable.} SequenceLayers gives you streaming \textit{for free}, in a production-friendly way. To achieve this, every streamable layer that has dependencies over time implements an explicit state, plus a step method that evolves that state (\Cref{sec:sequence-layer-api}).
    \item \textbf{Correct.} SequenceLayers is \textit{correct by default}, making entire classes of bugs impossible, e.g., those due to masking, upsampling, downsampling, causality, and padding-invariance. This comes from enforcing layer vs.\ step equivalence (\Cref{sec:sequence-layer-contract}) and tying mask information to sequence data everywhere with \texttt{Sequence} objects (\Cref{sec:sl-sequence}).
    \item \textbf{Composable.} SequenceLayers has an easy-to-understand \textit{declarative} API that enforces guarantees, enabling sequence models with concise definitions that read like block diagrams. These compositions are immediately streamable and expose aggregate properties like overall output ratio, latency, receptive field size, and state (\Cref{sec:combinators}).
\end{enumerate}

An example definition of a Transformer layer is shown in \Cref{fig:flax-vs-sl}. Due to the built-in \texttt{DotProductSelfAttention} layer's implementation of state (KV cache) and step, plus the automatic state wrapping and unwrapping performed by the \texttt{Serial} and \texttt{Residual} combinators, the aggregate layer exposes \texttt{get\_initial\_state()} and \texttt{step()} methods, allowing a sampling loop to be written with no further work.

\begin{figure*}[ht]
\begin{tcbitemize}[raster columns=2,
    raster force size=false,
    raster column 1/.style={width=0.5\textwidth},
    raster column 2/.style={width=0.5\textwidth},
    raster equal height,colframe=black!50,colback=white,colbacktitle=black!50,fonttitle=\rmfamily,left=1pt,right=1pt,top=1pt,bottom=1pt,
]
\tcbitem[title={Flax Transformer block}]
\begin{tcblisting}{
    listing engine=minted,
    minted style=friendly,
    minted language=python,
    listing only,
    left=0pt,right=0pt,top=0pt,bottom=0pt,
    boxrule=0pt,
    colback=white,
}

def setup(self) -> None:
  """Create submodules."""
  self.pre_attention_norm = ...
  self.attn = ...
  self.post_attention_norm = ...
  self.attn_dropout = ...
  self.pre_ffw_norm = ...
  self.mlp = ...
  self.post_ffw_norm = ...
  self.ffw_dropout = ...

def __call__(
  self,
  x: Float[ArrayT, 'B L D'],
  input_mask: Bool[ArrayT, 'B L'],
  attn_cache: AttentionKVCache | None,
  training: bool,
) -> tuple[Float[ArrayT, 'B L D'], 
           AttentionKVCache | None]:
  # Self attention.
  h = self.pre_attention_norm(x)
  attn, new_cache = self.attn(
    h, attn_cache=attn_cache)
  attn = self.post_attention_norm(attn)
  attn = self.attn_dropout(
    attn, training=training)
  x = x + attn
  
  # Gated GeLU FFN.
  norm_x = self.pre_ffw_norm(x)
  ffw_x = self.mlp(norm_x, input_mask)
  ffw_x = self.post_ffw_norm(ffw_x)
  ffw_x = self.ffw_dropout(
    ffw_x, training=training)
  output = x + ffw_x
  return output, new_cache

y = block(x, mask)
\end{tcblisting}

\tcbitem[title={SequenceLayer Transformer block}]
\begin{tcblisting}{
    listing engine=minted,
    minted style=friendly,
    minted language=python,
    listing only,
    left=0pt,right=0pt,top=0pt,bottom=0pt,
    boxrule=0pt,
    colback=white,
}
import sequence_layers.jax as sl

block = sl.Serial.Config([

  # Self attention.
  sl.Residual.Config([
    sl.RMSNormalization.Config(
      name='pre_norm'),
    sl.DotProductSelfAttention.Config(
      num_heads=16, 
      units_per_head=64,
      # Global causal attention.
      max_past_horizon=-1,  
      max_future_horizon=0,      
      name='self_attention'),
    sl.DenseShaped.Config(
      [d_model], name='out_proj'),
    sl.RMSNormalization.Config(
      name='post_norm'),
    sl.Dropout.Config(dropout_rate),
  ], name='attention_block'),
  
  # Gated GeLU FFN.
  sl.Residual.Config([
    sl.RMSNormalization.Config(
      name='pre_norm'),
    sl.Dense.Config(
      4 * d_model, name='dense1'),
    sl.GatedUnit.Config(jax.nn.gelu, None),
    sl.Dense.Config(d_model, name='dense2'),
    sl.RMSNormalization.Config(
      name='post_norm'),
    sl.Dropout.Config(dropout_rate),
  ], name='ffn_block')
  
], name='transformer_block').make()

y = block.layer(x, training=False)

\end{tcblisting}

\end{tcbitemize}
  \caption{Imperative forward pass definition of a Transformer block in Flax, versus a declarative definition in SequenceLayers for JAX. Unlike other declarative libraries, this SequenceLayer definition gives you a stateful \texttt{step()} method out of the box, due to the contracts implemented by each component layer and combinator. Imperative \texttt{setup()} methods and implementation overrides remain available, allowing the mix-and-match of paradigms.}
  \label{fig:flax-vs-sl}
\end{figure*}

\subsection{Related Work}

We take inspiration for our declarative and composable API from Trax \citep{trax2019}, which demonstrates the effectiveness of serial and parallel combinators (\Cref{sec:combinators}). We simplify their approach by removing the data stack, and add the additional capability of streamability.

While existing libraries such as Keras \citep{chollet2018keras}, Trax \citep{trax2019} and others promote a declarative and compositional API, to the best of our knowledge no other layer library or framework leverage explicit state in a uniform API in order to enable streaming of compositions of layers.

Unlike other fully featured deep learning frameworks such as T5X \citep{roberts2023t5x}, HuggingFace Transformers \citep{wolf2019huggingface}, and Lingvo \citep{shen2019lingvomodularscalableframework}, we intentionally limit the scope of SequenceLayers to \textit{layers}. We view training objectives, data loading, and optimization as out-of-scope. This keeps SequenceLayers lightweight and adaptable to one's model framework of choice.\\

\textbf{SequenceLayers is a design, not an implementation.} Although it is implemented for JAX \citep{frostig2018jax} and TensorFlow \citep{abadi2016tensorflow}, its design is agnostic to the underlying neural network framework and could be ported to e.g., PyTorch \citep{paszke2019pytorch} or MLX \citep{mlx2023}.

\section{Design: Core Features}

\label{sec:sl-sequence}
SequenceLayers consume and produce \texttt{Sequence} objects, which are lightweight, differentiable, pytree\footnote{\url{https://docs.jax.dev/en/latest/pytrees.html}} dataclasses that pair sequence \texttt{values} of shape \texttt{[batch, time, ...]} with a boolean \texttt{mask} of shape \texttt{[batch, time]} indicating valid positions in the batch. The \texttt{...} dimensions following the \texttt{[batch, time]} dimensions are the \textbf{channel shape} of the \texttt{Sequence}. A \textbf{spec} is simply a \texttt{jax.ShapeDtypeStruct} or \texttt{tf.TensorSpec} indicating shape, dtype, and sharding information.

\texttt{Sequence}s are array-like objects, in that they implement properties of \texttt{jnp.ndarray} or \texttt{np.ndarray} like \texttt{shape}, \texttt{dtype}, and \texttt{ndim}. However, \texttt{Sequence} does not implement the array protocol\footnote{\url{https://numpy.org/doc/stable/reference/arrays.interface.html}} to avoid accidentally using a \texttt{Sequence}'s values without properly respecting the mask.

A \texttt{Sequence} is said to be \textbf{masked} if for all batch positions $b$ and time positions $t$, \texttt{mask[b, t]} is \texttt{False} implies that \texttt{values[b, t, ...]} equals zero. The helper method \texttt{mask\_invalid()} returns a new \texttt{Sequence} with this property. Whether a sequence is known to be masked is represented by a marker type \texttt{MaskedSequence} which is a subclass of \texttt{Sequence} with \texttt{mask\_invalid()} replaced with a no-op. When critical for performance, one can declare a \texttt{MaskedSequence} which promises that invalid positions (\texttt{mask=False}) are already zeroed out (\Cref{fig:sl-sequence}).

\texttt{Sequence}s have convenience methods to streamline creation, manipulation, and information gathering (e.g., \texttt{from\_values(), from\_lengths(), lengths(), pad\_time()}), as well as to support usual shorthands for slicing (\texttt{seq[..., a:b]}).

\begin{figure*}[ht]
\begin{tcbitemize}[raster columns=2,
    raster force size=false,
    raster column 1/.style={width=0.55\textwidth},
    raster column 2/.style={width=0.45\textwidth},
    raster equal height,colframe=black!50,colback=white,colbacktitle=black!50,fonttitle=\rmfamily,left=1pt,right=1pt,top=1pt,bottom=1pt,
]
\tcbitem[title={Code}]
\begin{tcblisting}{
    listing engine=minted,
    minted style=friendly,
    minted language=python,
    listing only,
    left=0pt,right=0pt,top=0pt,bottom=0pt,
    boxrule=0pt,
    colback=white,
}
x = sl.Sequence(
    jnp.ones((2, 3)),
    jnp.asarray([[True, True, False],
                 [True, False, False]])
)
# Helper that zeroes invalid positions:
x = x.mask_invalid()  # Returns sl.MaskedSequence.
x = x.mask_invalid()  # No-op.
\end{tcblisting}
\tcbitem[title={Value of \texttt{x}:},colframe=white,coltitle=black,colbacktitle=white]
\begin{tcblisting}{
    listing engine=minted,
    minted style=friendly,
    minted language=python,
    listing only,
    left=0pt,right=0pt,top=0pt,bottom=0pt,
    boxrule=0pt,
}
MaskedSequence(
    values=Array([[1., 1., 0.],
                  [1., 0., 0.]],
                  dtype=float32),
    mask=Array([[ True,  True, False],
                [ True, False, False]],
                dtype=bool)
)
    \end{tcblisting}
\end{tcbitemize}
  \caption{Example demonstrating the \texttt{Sequence} and  \texttt{MaskedSequence} primitives.}
  \label{fig:sl-sequence}
\end{figure*}

\subsection{\emph{Streamable:} The \texttt{SequenceLayer} Type}
\label{sec:sequence-layer-api}

\texttt{SequenceLayer} is a Python class which defines the basic API and functionality required to achieve the goals of the library.

The fundamental methods of the API are:
\begin{itemize}
\item \textbf{\texttt{layer: Sequence -> Sequence}}: Process a sequence \texttt{x} layer-wise and return a new sequence \texttt{y}.
\item \textbf{\texttt{get\_initial\_state: State}}: Returns a pytree of state arrays for step-wise execution of the layer.
\item \textbf{\texttt{step: (Sequence, State) -> (Sequence, State)}}: Processes one block of inputs \texttt{x} and produces a block of outputs \texttt{y}.
\end{itemize}

A \texttt{SequenceLayer} that is steppable must produce identical outputs for an input sequence regardless of whether the \texttt{layer} or \texttt{step} API is used (Figure \ref{fig:layer-vs-step}).

\begin{figure*}[h]
\begin{tcbitemize}[raster columns=1,
    raster force size=false,
    raster column 1/.style={width=1.0\textwidth},
    raster equal height,colframe=black!50,colback=white,colbacktitle=black!50,fonttitle=\rmfamily,left=1pt,right=1pt,top=1pt,bottom=1pt,
]
\tcbitem[title={Code}]
\begin{tcblisting}{
    listing engine=minted,
    minted style=friendly,
    minted language=python,
    listing only,
    left=0pt,right=0pt,top=0pt,bottom=0pt,
    boxrule=0pt,
    colback=white,
}
# Define a test model and input.
model = sl.Conv1D.Config(filters=5, kernel_size=3, padding='causal').make()
x = sl.Sequence.from_values(
  jax.random.uniform(key, (b, t, c))
)

# Process x layer-wise.
y = model.layer(x, training=True)

# Process x step-wise in blocks of size 2.
state = model.get_initial_state(b, x.channel_spec, training=True)
y0, state = model.step(x[:, 0:2], state, training=True)
y1, state = model.step(x[:, 2:4], state, training=True)
y2, state = model.step(x[:, 4:6], state, training=True)
...

y_step = sl.Sequence.concatenate_sequences([y0, y1, y2, ...])

# Layer-wise and step-wise outputs must be equivalent.
np.testing.assert_array_equal(y.values, y_step.values)
np.testing.assert_array_equal(y.mask, y_step.mask)

\end{tcblisting}

\end{tcbitemize}
  \caption{Example demonstrating layer-wise and step-wise execution of a \texttt{SequenceLayer}.}
  \label{fig:layer-vs-step}
\end{figure*}

\subsubsection{Output Ratio and Block Size}
\label{sec:output-ratio-block-size}

A \texttt{SequenceLayer} can downsample or upsample its input. However, the ratio of output timesteps to input timesteps must be a constant, due to the requirement in compiled JAX or TensorFlow programs that all functions must have fixed shapes and types. This requirement implies that the output shape cannot vary across invocations of any compiled \texttt{SequenceLayer} function.

The \textbf{output ratio} of a \texttt{SequenceLayer} is a constant ratio (represented as a \texttt{fractions.Fraction}) between the number of output timesteps and input timesteps to a layer. For example:

\begin{center}
\begin{tabular}{|l|l|l|}
 \hline
 Layer & Config & Output Ratio \\
 \hline
 \texttt{Dense} & --- & \texttt{Fraction(1, 1)} \\
 \texttt{Conv1D} & \texttt{stride = 1} & \texttt{Fraction(1, 1)} \\
 \texttt{Conv1D} & \texttt{stride = 2} & \texttt{Fraction(1, 2)} \\
 \texttt{Conv1DTranspose} & \texttt{stride = 1} & \texttt{Fraction(1, 1)} \\
 \texttt{Conv1DTranspose} & \texttt{stride = 2} & \texttt{Fraction(2, )} \\
 \texttt{Downsample1D} & \texttt{rate = 2} & \texttt{Fraction(1, 2)} \\
 \texttt{Upsample1D} & \texttt{rate = 2} & \texttt{Fraction(2, 1)} \\
 \texttt{DotProductSelfAttention} & --- & \texttt{Fraction(1, 1)} \\
 \hline
\end{tabular}
\end{center}

Since the input and output shapes of compiled JAX and TensorFlow programs must be constant, a \texttt{SequenceLayer} \texttt{step} must always produce at least one output timestep for a group of input timesteps. Since the number of input and output timesteps must be integral and each layer has a constant output ratio, each layer therefore has a \textbf{block size} which the total number of input timesteps to \texttt{step} must be divisible by in order to produce an integral number of outputs. This is necessarily at least $\nicefrac{1}{\texttt{output\_ratio}}$, but may be larger (for example, if a layer internally downsamples by 2x and upsamples by 2x, its output ratio would be 1 but its block size would be 2).

\subsubsection{State}

As shown in \Cref{fig:layer-vs-step}, \textbf{\texttt{State}} is a key part of the step-wise execution of the layer. An initial state is returned from the \texttt{get\_initial\_state} method, then the \texttt{state} is repeatedly updated by the \texttt{step} method to produce the next output and state for an input block and previous \texttt{state}.

The \texttt{State} type is any pytree of fixed shape / type arrays representing layer state that evolves over repeated inference calls, such as the KV cache for an attention layer, a buffer for a convolution, or a cell/hidden state for an LSTM.

All SequenceLayer step-wise state is represented via explicit arrays passed into and out of layer methods. No state is stored within the layer (for example via Flax variables\footnote{\url{https://flax-linen.readthedocs.io/en/latest/guides/flax_fundamentals/state_params.html}}). This makes SequenceLayers well suited to the JAX pure functional programming model, despite being implemented in an object oriented manner.

To ensure that \texttt{State}s are passed to the correct sublayer of an aggregate SequenceLayer, combinators like \texttt{Serial} (\Cref{sec:combinators-serial}) wrap and unwrap them into container pytrees in a well-defined and ordered way (e.g., \texttt{Serial} takes and returns a \texttt{State} which is a tuple of its sublayers' \texttt{State}s).

Note that from the perspective of the computation graph compiler (e.g., XLA; \citealp{sabne2020xla}), there is no difference between data passed via these containers versus the main \texttt{Sequence} layer inputs and outputs; rather, these semantic conventions are for the benefit of SequenceLayer users and developers. Hence, if the shapes and dtypes of a \texttt{State}'s leaves vary across invocations, this may lead to unnecessary recompilation, just as differently shaped \texttt{Sequence} inputs would.

\subsubsection{Constants}
\label{sec:constants}

\textbf{\texttt{Constants}} is an optional mutable dictionary of pytrees (\texttt{MutableMapping[str, pytree]}) representing auxiliary inputs to the \texttt{layer}, \texttt{step}, and \texttt{get\_initial\_state} methods. 
These could be conditioning vectors / \texttt{Sequence}s which also become part of the computation graph (e.g., during cross-attention), or custom user-defined effects on control flow (e.g., \texttt{`causal':\ False}). The Constants dictionary is mutable and its contents are read in the same order that \texttt{Sequence} inputs/outputs are propagated within a \texttt{SequenceLayer} (e.g., first to last over the layers of a \texttt{Serial}).

For simplicity, constants are available to all layers via the flat namespace of the string keys to the dictionary. Combinators like \texttt{Serial} and \texttt{Parallel} (\Cref{sec:combinators}) broadcast constants to all sub-layers. This choice comes with the possibility of namespace clashes, but simplifies the API for the common case of providing cross attention sources and conditioning arrays to specific layers. We considered a more complex routing scheme, but decided that it was likely that users could make the keys unique using assumptions about their specific model, which is always easier than building a general-purpose solution that will be infrequently used.

With \texttt{Constants}, the API in \Cref{sec:sequence-layer-api} becomes: 
\begin{itemize}
\item \textbf{\texttt{layer: (Sequence, Constants) -> Sequence}}: Process a sequence \texttt{x} layer-wise and return a new sequence \texttt{y}.
\item \textbf{\texttt{get\_initial\_state: Constants -> State}}: Returns a pytree of state arrays for step-wise execution of the layer.
\item \textbf{\texttt{step: (Sequence, State, Constants) -> (Sequence, State)}}: Processes a block of inputs \texttt{x} and produces a block of outputs \texttt{y}.
\end{itemize}

\subsubsection{Latency and Lookahead}

A \texttt{SequenceLayer} can introduce lookahead or delay in its output. A \texttt{Delay} layer delays its input by $N$ timesteps, while a \texttt{Lookahead} layer drops $N$ inputs to jump forward in the input. Layers like \texttt{DotProductSelfAttention} and \texttt{Conv1D} support lookback and lookahead as well without shifting the sequence time alignment.

Whatever the behavior of the \texttt{layer} method, if the layer supports stepping, the \texttt{step} method must produce identical results to be considered a valid \texttt{SequenceLayer}. Since input arrives in a stream of multiples of \texttt{block\_size} timesteps, outputs from \texttt{step} may have to wait until enough inputs have arrived to perform the same function as \texttt{layer}.

To implement this waiting, a layer may output invalid (\texttt{mask = False}) timesteps, waiting for more inputs before it produces a valid timestep. The number of output timesteps the caller must discard before expecting the first valid timestep from a layer is the layer's \textbf{output latency}. This is programmatically available via the \texttt{output\_latency} property of the layer.

As a consequence of the output latency (which may entail internal buffering within the layer \texttt{State}), to produce all of the valid outputs from the layer it may be necessary to feed invalid inputs to the layer to ``flush'' it. The number of inputs required to flush the layer is the layer's \textbf{input latency}. This is programmatically available via the \texttt{input\_latency} property of the layer. 

The input and output latency are often the same, but may differ. For example, when a layer upsamples or downsamples its input with lookahead, the input and output latency will be different since the time dimension has different scales in the input sequence and output sequence.

With these definitions in place, we can now correct an oversimplification in \Cref{fig:layer-vs-step}. \Cref{fig:lookahead-layer-step-equivalence} demonstrates the full logic required to achieve layer/step equivalence when the layers in use have lookahead or delay.

\begin{figure*}[h]
\begin{tcbitemize}[raster columns=1,
    raster force size=false,
    raster column 1/.style={width=1.0\textwidth},
    raster equal height,colframe=black!50,colback=white,colbacktitle=black!50,fonttitle=\rmfamily,left=1pt,right=1pt,top=1pt,bottom=1pt,
]
\tcbitem[title={Code}]
\begin{tcblisting}{
    listing engine=minted,
    minted style=friendly,
    minted language=python,
    listing only,
    left=0pt,right=0pt,top=0pt,bottom=0pt,
    boxrule=0pt,
    colback=white,
}
# A convolution layer with 4 steps of lookahead.
model = sl.Conv1D.Config(filters=3, kernel_size=5, padding='reverse_causal').make()
x = sl.Sequence.from_values(
  jax.random.uniform(key, (b, t, c))
)

# Process x layer-wise.
y = model.layer(x, training=True)

assert model.input_latency == 4
assert model.output_latency == 4

# "Flush" the layer with input_latency invalid timesteps.
x_padded = x.pad_time(0, model.input_latency, valid=False)

y_step, _, _ = sl_utils.step_by_step_dynamic(model, x_padded, training=False)

# Ignore the first output_latency timesteps since they are invalid (mask = False, since the 
# first valid output cannot be produced until 4 lookahead timesteps have arrived).
y_step = y_step[:, model.output_latency:]

# Layer-wise and step-wise outputs are equivalent after accounting for latency.
np.testing.assert_array_equal(y.values, y_step.values)
np.testing.assert_array_equal(y.mask, y_step.mask)

\end{tcblisting}

\end{tcbitemize}
  \caption{Demonstration of layer / step equivalence when the \texttt{SequenceLayer} has lookahead.}
  \label{fig:lookahead-layer-step-equivalence}
\end{figure*}

\subsubsection{Emits}

Since the \texttt{layer} and \texttt{step} APIs return a single \texttt{Sequence} output, there is no easy way for layers to produce auxiliary debugging output.

To support this use case, we introduce additional APIs that return \textbf{\texttt{Emits}}. An \texttt{Emit} is a layer-defined pytree of arrays or sequences. 

The calculation of \texttt{Emits} may entail extra work within a layer. While optimizing compilers like XLA \citep{sabne2020xla} can automatically prune unused computation, we would like to avoid the need for compiling this code in the first place, or the execution of the code in eager mode when the auxiliary outputs will not be used. 

To this end, we provide additional optional APIs for computing \texttt{layer} and \texttt{step} with auxiliary outputs: 
\begin{itemize}
\item \textbf{\texttt{layer\_with\_emits: (Sequence, Constants) -> (Sequence, Emits)}}: Process a sequence \texttt{x} layer-wise and return a new sequence \texttt{y} and auxiliary emits.
\item \textbf{\texttt{step\_with\_emits: (Sequence, State, Constants) -> (Sequence, State, Emits)}}: Processes a block of inputs \texttt{x} and produces a block of outputs \texttt{y} and auxiliary emits.
\end{itemize}

Additionally, we provide a convenience \texttt{Emitting} sub-class of \texttt{SequenceLayer} that implements \texttt{layer} and \texttt{step} in terms of \texttt{layer\_with\_emits} and \texttt{step\_with\_emits}, as well as an \texttt{Emit} \texttt{SequenceLayer} that simply emits the input to the layer as an emit (for easily ``tapping'' into the intermediate sequences traversing a stack of layers).

\subsubsection{Receptive Field}
Having access to accurately computed layer and architecture receptive fields facilitates network design when precise control over model causality is desired. To simplify the calculation of receptive fields for any architecture, we developed the \texttt{receptive\_field} property. This utility determines the effective range of input time steps that affect a single output time step. When dealing with receptive fields, several key challenges arise:

\begin{itemize}

\item \textbf{Layers with \texttt{output\_ratio != 1}}: We define the receptive field as the \texttt{(start, end)} tuple that describes the input step range \texttt{[t\_i + start, t\_i + end]} that affects the output time step \texttt{t\_o}, where \texttt{t\_i = t\_o // output\_ratio}.

\item \textbf{Layers with alternating receptive fields every n steps}: To handle this complexity, we keep track of a step-specific receptive field map via the \texttt{receptive\_field\_per\_step} helper property. The overall \texttt{receptive\_field} is the union of the step-specific receptive fields and describes the maximal temporal window of influence for the layer. \texttt{receptive\_field\_per\_step} is used by combinator layers such as \texttt{Serial} to compute the overall \texttt{receptive\_field\_per\_step} for compositions of layers.

\item \textbf{Layers with infinite or no receptive field:} The implementation is robust to special cases, representing infinite receptive fields (e.g., \texttt{LSTM} has \texttt{(-inf, 0)} receptive field) and layers with holes in their receptive field (e.g., \texttt{Conv1DTranspose} with \texttt{stride > kernel\_size} has a \texttt{None} receptive field for some time steps).

\end{itemize}

See \Cref{fig:receptive-field} for examples of the \texttt{receptive\_field} property in action.

\begin{figure*}[ht]
\begin{tcbitemize}[raster columns=1,
    raster force size=false,
    raster column 1/.style={width=1.0\textwidth},
    raster equal height,colframe=black!50,colback=white,colbacktitle=black!50,fonttitle=\rmfamily,left=1pt,right=1pt,top=1pt,bottom=1pt,
]
\tcbitem[title={Code}]
\begin{tcblisting}{
    listing engine=minted,
    minted style=friendly,
    minted language=python,
    listing only,
    left=0pt,right=0pt,top=0pt,bottom=0pt,
    boxrule=0pt,
    colback=white,
}
# Convolution layer with causal padding.
model = sl.Conv1D.Config(filters=3, kernel_size=5, padding='causal').make()
assert model.receptive_field == (-4, 0)

# Convolution layer with reverse_causal padding.
model = sl.Conv1D.Config(filters=3, kernel_size=5, padding='reverse_causal').make()
assert model.receptive_field == (0, 4)

# Convolution layer with same padding.
model = sl.Conv1D.Config(filters=3, kernel_size=5, padding='same').make()
assert model.receptive_field == (-2, 2)

# Stack of convolution layers with same padding.
model = sl.Serial.Config([Conv1D.Config(filters=3, kernel_size=5, padding='same')] * 4).make()
assert model.receptive_field == (-8, 8)

# LSTM.
model = sl.LSTM.Config(units=3).make()
assert model.receptive_field == (-np.inf, 0)

# Transposed convolution with kernel_size < stride.
model = sl.Conv1DTranspose.Config(filters=1, kernel_size=1, strides=2, padding='same').make()
assert model.receptive_field_per_step == {0: (0, 0), 1: None}
assert model.receptive_field == (0, 0)

# Mixed downsampling + upsampling layers.
model = sl.Serial.Config(
    [
        sl.Conv1D.Config(filters=1, kernel_size=5, strides=2, padding='same'),
        sl.Conv1DTranspose.Config(filters=1, kernel_size=6, strides=4, padding='same'),
    ]
).make()
assert model.receptive_field_per_step == {0: (-4, 2), 1: (-2, 2), 2: (-2, 2), 3: (-2, 4)}
assert model.receptive_field == (-4, 3)
\end{tcblisting}

\end{tcbitemize}
  \caption{Demonstration of the \texttt{receptive\_field} property for various layers.}
  \label{fig:receptive-field}
\end{figure*}

\subsection{\emph{Correct:} The \texttt{SequenceLayer} Contract}
\label{sec:sequence-layer-contract}

In this section, we introduce the \textbf{SequenceLayer contract}, which is the set of requirements a \texttt{SequenceLayer} must implement to be considered correct.

\begin{itemize}
    \item \textbf{Layer-wise and step-wise equivalence}: If step-wise operation is supported, the \texttt{layer} and \texttt{step} methods must produce identical results when fed identical data and starting state (slicing the data into blocks of any multiple of \texttt{block\_size} timesteps. See Figure \ref{fig:layer-vs-step} for a code example. Stateful stochastic layers (e.g., \texttt{Dropout}) should obey this property when the starting RNG state is equivalent.
    \item \textbf{Padding and batching invariance}: The \texttt{layer} and \texttt{step} methods must produce identical results when fed identical data with differing amounts of end padding, or when the position of examples in a batch is shuffled. For the common use-case of batching contiguous sequences of mixed lengths together, the lengths of other sequences in the batch or the position in the batch should have no bearing on the calculation performed by the layer.

\textit{Corollary}: Padding values must not affect the calculation of non-padding values.

\textit{Note}: Padding invariance is currently only required for end padding. Start padding or interior padding (for non-contiguous sequences) does affect the behavior of calculations. This may change in the future.
    
    \item \textbf{Masked inputs and outputs}: For an input \texttt{Sequence} provided to a \texttt{SequenceLayer} with \texttt{values} (\texttt{[b, t, ...]}) and \texttt{mask} (\texttt{[b, t]}), layers must not assume \texttt{values} is masked. If the computation performed by the layer requires masked inputs (e.g., it mixes information across timesteps), then the layer must mask the input sequence before use. The layer may return either a \texttt{Sequence} or a \texttt{MaskedSequence}.
    
\end{itemize}

Each \texttt{SequenceLayer} included with the library has unit tests that it obeys this contract. You can test that your own layers obey this contract using the \texttt{verify\_contract} test utility provided with the library. \texttt{verify\_contract} performs the following compliance tests:
\begin{itemize}
    \item \texttt{layer} and \texttt{step} (taking steps of \texttt{block\_size} timesteps) produce identical outputs (up to floating point tolerance) on the same input sequence.
    \item \texttt{layer} and \texttt{step} produce identical gradients with respect to both their parameters and inputs.
    \item A step-wise application with $2 \times \texttt{block\_size}$ produces identical outputs to the layer-wise output.
    \item The layer's behavior is consistent with its metadata  (\texttt{get\_output\_spec}, \texttt{input\_latency}, \texttt{output\_latency}, \texttt{output\_ratio}, \texttt{block\_size}).
    \item The \texttt{receptive\_field} property matches a gradient-based calculation of the receptive field.
    \item The layer is \textbf{batching invariant}, by inserting additional invalid batch items and verifying equivalent output.
    \item The layer is \textbf{padding invariant}, by replacing all invalid timesteps with \texttt{NaN}s or large-valued integers and verifying the outputs for valid timesteps are unchanged.
\end{itemize}

These tests are typically invoked with randomly generated input sequences and layer parameters (taking care to avoid zero-initialization for bias-like parameters).

Additionally, for ease of streaming deployment nearly all layers in the library have unit tests that their \texttt{step} function can be exported as a TensorFlow saved model and converted to LiteRT for mobile deployment.

\subsection{\emph{Composable:} Declarative API and Combinators}
\label{sec:combinators}

Since every \texttt{SequenceLayer} has a uniform API for layer-wise and step-wise processing, it is easy to build \textbf{combinators} or \texttt{SequenceLayer}s that are compositions of other \texttt{SequenceLayer}s.

\subsubsection{The \texttt{Serial} Combinator}
\label{sec:combinators-serial}

The simplest combinator is the \texttt{Serial} combinator (\Cref{fig:combinators-serial}), which simply executes a list of \texttt{SequenceLayers} serially.

\begin{figure*}[h]
\begin{tcbitemize}[raster columns=1,
    raster force size=false,
    raster column 1/.style={width=1.0\textwidth},
    raster equal height,colframe=black!50,colback=white,colbacktitle=black!50,fonttitle=\rmfamily,left=1pt,right=1pt,top=1pt,bottom=1pt,
]
\tcbitem[title={Code}]
\begin{tcblisting}{
    listing engine=minted,
    minted style=friendly,
    minted language=python,
    listing only,
    left=0pt,right=0pt,top=0pt,bottom=0pt,
    boxrule=0pt,
    colback=white,
}
# Define a serial of 2 causal convolutions.
model = sl.Serial.Config([
  sl.Conv1D.Config(filters=5, kernel_size=3, stride=2, padding='causal'),
  sl.Conv1D.Config(filters=8, kernel_size=5, stride=3, padding='causal'),
]).make()

x = sl.Sequence.from_values(
  jax.random.uniform(key, (b, t, c))
)

# Applies Conv-Conv serially layer-wise and step-wise.
y = model.layer(x, training=True)
y_step, _, _  = sl_utils.step_by_step_dynamic(l, x, training=True)

# Layer-wise and step-wise outputs must be equivalent.
np.testing.assert_array_equal(y.values, y_step.values)
np.testing.assert_array_equal(y.mask, y_step.mask)

# A stride 2 and stride 3 convolution decimates the sequence by 6x.
assert model.output_ratio == fractions.Fraction(1, 6)
assert model.block_size == 6
assert y.shape == (b, t // 6, 8)
\end{tcblisting}

\end{tcbitemize}
  \caption{The \texttt{Serial} combinator.}
  \label{fig:combinators-serial}
\end{figure*}

\subsubsection{The \texttt{Parallel} Combinator}
\label{sec:combinators-parallel}

The \texttt{Parallel} combinator enables processing an input sequence in parallel by two or more \texttt{SequenceLayer}s, combining the result at the end according to a fixed number of broadcasted combination strategies (for example, stacking channels, concatenating on the final channel axis, adding channels, averaging across channels).

\subsubsection{The \texttt{Residual} Combinator}
\label{sec:combinators-residual}

The \texttt{Residual} combinator simply transforms a \texttt{SequenceLayer} implementing a function $F(x)$ into a residual function $F(x) + x$. Due to the critical importance of residual functions to deep learning \citep{he2015resnet}, for readability this deserves a dedicated combinator even though it can be implemented with \texttt{Parallel}.

\subsubsection{The \texttt{Repeat} Combinator}
\label{sec:combinators-repeat}

The \texttt{Repeat} combinator (\Cref{fig:combinators-repeat}) repeats a specified \texttt{SequenceLayer} a certain number of times using a control flow primitive such as \texttt{jax.lax.scan}. A different set of layer parameters is used for each iteration of the loop. 

Using a loop to process the input implies the shape and type of the input is unchanged throughout all iterations of the loop and the calculation performed across all iterations only differs in the inputs and parameters. This is useful for reducing compilation time when compiling large models, since the optimizing compiler only has to compile the function once. This comes with the downsides of not being able to optimize across iterations of the loop. The \texttt{unroll\_layer} and \texttt{unroll\_step} options can be used to unroll either the \texttt{layer} or \texttt{step} operations respectively to enable cross-iteration optimization in either program separately. A \texttt{remat} option wraps each iteration of the loop in a gradient checkpoint, so that peak memory usage during backpropagation is reduced.

\begin{figure*}[h]
\begin{tcbitemize}[raster columns=1,
    raster force size=false,
    raster column 1/.style={width=1.0\textwidth},
    raster equal height,colframe=black!50,colback=white,colbacktitle=black!50,fonttitle=\rmfamily,left=1pt,right=1pt,top=1pt,bottom=1pt,
]
\tcbitem[title={Code}]
\begin{tcblisting}{
    listing engine=minted,
    minted style=friendly,
    minted language=python,
    listing only,
    left=0pt,right=0pt,top=0pt,bottom=0pt,
    boxrule=0pt,
    colback=white,
}
# Repeat TransformerBlock num_blocks times.
model = sl.Repeat.Config(
  TransformerBlock(d_model, ...),
  num_repeats=num_blocks,
  # Checkpoint gradients to reduce memory usage in backpropagation.
  remat=True,
).make()
\end{tcblisting}

\end{tcbitemize}
  \caption{The \texttt{Repeat} combinator.}
  \label{fig:combinators-repeat}
\end{figure*}

\subsubsection{The \texttt{Bidirectional} Combinator}
\label{sec:combinators-bidirectional}

The \texttt{Bidirectional} combinator processes its input in the forward direction with a \texttt{forward} layer, and in the backward direction with a \texttt{backward} layer, and then combines the resulting forward and backward sequences. Since this entails reversing the input sequence, it is not steppable. This is effectively a generalization of bidirectional RNNs \citep{bahdanau2016neuralmachinetranslationjointly} to any forward and backward network.

\subsubsection{The \texttt{Blockwise} Combinator}
\label{sec:combinators-blockwise}

The \texttt{Blockwise} combinator (\Cref{fig:combinators-blockwise}) is a useful tool for adjusting the block size of any \texttt{SequenceLayer} and automatically re-implementing the \texttt{layer} method in terms of steps of a chosen \texttt{block\_size}. This enables adjusting the native streaming block size of a layer without changing any execution code (as long as the execution code uses the \texttt{block\_size} property of the layer). Additionally, it enables limiting the peak memory usage of the \texttt{layer} method by splitting the input sequence into blocks of size \texttt{block\_size} for processing using the \texttt{step} method within \texttt{layer}.

\begin{figure*}[h]
\begin{tcbitemize}[raster columns=1,
    raster force size=false,
    raster column 1/.style={width=1.0\textwidth},
    raster equal height,colframe=black!50,colback=white,colbacktitle=black!50,fonttitle=\rmfamily,left=1pt,right=1pt,top=1pt,bottom=1pt,
]
\tcbitem[title={Code}]
\begin{tcblisting}{
    listing engine=minted,
    minted style=friendly,
    minted language=python,
    listing only,
    left=0pt,right=0pt,top=0pt,bottom=0pt,
    boxrule=0pt,
    colback=white,
}
# Process the input sequence 1024 steps at a time.
model = sl.Blockwise.Config(
  TransformerConfig(d_model, ...),
  block_size=1024,
).make()

assert model.block_size == 1024

# One million timesteps of input, too large to process given our memory constraints.
x = sl.Sequence.from_values(
  jax.random.normal(key, (b, 1000000, d_model))
)

# Layer-wise application of 1 million timesteps streams 1024 steps at a time,
# reducing peak memory usage.
y = model.layer(x, training=True)

# Step-wise application of 1 million timesteps streams 1024 steps at a time, 
# increasing throughput versus the default of 1 timestep.
y_step, _, _  = sl_utils.step_by_step_dynamic(l, x, training=True)
\end{tcblisting}

\end{tcbitemize}
  \caption{The \texttt{Blockwise} combinator.}
  \label{fig:combinators-blockwise}
\end{figure*}

\section{Design: Other Features and Choices}

\subsection{The Tradeoffs of Modularity}

Modularity and abstraction are the cornerstone of not just computer science and engineering, but all forms of engineering. It is impossible to build maintainable large systems or structures without some degree of modularity. Additionally, in large organizations modularity and division of responsibility enables teams crossing timezones to work together effectively to build large systems.

The key to success is to choose the right places to put the abstraction or module boundaries. In some situations, it will be  necessary to break an abstraction boundary in the name of efficiency. This may be a sign that a suboptimal boundary was chosen, or simply a sign that the system requirements have changed.

The overall SequenceLayers design is helpful when defining the abstraction boundaries of a system. 
It can be used to declare a priori that the only thing that matters about a specific sequence-processing block to the broader system it is used within is its underlying sequence-processing properties:

\begin{itemize}[topsep=1pt,itemsep=1pt]
    \item The shape and type of the input and output sequence.
    \item The block sizes of inputs that it can operate over.
    \item Whether the module decimates or upsamples its input.
    \item The latency (delay or lookahead) in sequence time (not wallclock time) induced by the module.
    \item The receptive field of the module; the causal relationship between inputs and outputs of the layer.
    \item The compute profile (the wallclock processing time for its execution, the size of its state arrays, the size of the model parameters, the number of accelerators it requires).
\end{itemize}

The exact form of the function and the details of its state can be left as a black box. This is powerful for decoupling architectural details (e.g., Transformer vs.\ RNN) from the role the architecture plays in a broader system (e.g., encode an input sequence non-causally for use by a causal decoder, parameterize the distribution over next-token probabilities given previous samples, predict the noise velocities at the current noise level as part of a diffusion process).

However, there may come a time when the modularity serves as a hindrance. For example, the trend of sharing KV caches across blocks in a Transformer model \citep{sun2024yoco} breaks the modularity of the blocks. While it's possible to build a new combinator for this specific style of state sharing, another approach is to simply redraw the abstraction boundary of the \texttt{SequenceLayer} to cover the entire Transformer architecture rather than describing each Transformer block as a composition of \text{SequenceLayer}s. However, from the perspective of consumers of the Transformer nothing has changed and they can continue to abstract the details of the Transformer via a \texttt{SequenceLayer}. In our view, the benefits accrued from modularity greatly outweigh the papercut of such a refactoring.

\subsection{Training Mode}

The core \texttt{SequenceLayer} API methods (\texttt{layer}, \texttt{get\_initial\_state}, \texttt{step}) all have a \textbf{required} \texttt{training} keyword argument, which at first glance is an aesthetic wart. In this section we discuss this design decision.

All deep learning layer libraries need to implement some method of changing behavior between training and test time. \texttt{Dropout} and \texttt{BatchNormalization} are canonical examples that produce this requirement.

Some frameworks opt to represent the specific behavior under control as a call-time parameter, for example Flax \citep{heek2024flax} provides a \texttt{deterministic} constructor and call time parameter to \texttt{Dropout}. This has the benefit of being explicit about the behavior being controlled. However, any parent layer calling the dropout layer will need a \texttt{training} parameter to know how to set this value. The end result is the same, but the interpretation of the \texttt{training} flag is left to parent layers instead of the leaf layers (which is useful when the decision is not one-size-fits-all).

Some frameworks such as Keras \citep{chollet2018keras} have a \texttt{training} keyword argument to the call method of certain layers (e.g., \texttt{keras.Dropout}) which defaults to \texttt{False}. However, users are not obligated to recursively pass the \texttt{training} parameter throughout the call chain of their model. Instead, the \texttt{keras.Layer.\_\_call\_\_} method automatically searches upward in the call stack for any manually specified \texttt{training} argument, and defaults to \texttt{False} otherwise. While this is a nice way of avoiding threading \texttt{training} recursively through all layer calls, it comes with significant implementation complexity (Keras owns the \texttt{\_\_call\_\_} method) which contributes to the feeling of Keras being a ``framework'' instead of a library. Our goal is to make SequenceLayers feel like a simple Python library. We therefore discarded this solution.

Some frameworks use a global variable to indicate the training phase (e.g., ``\texttt{learning\_phase}'' in TensorFlow Keras). We also discard this solution since global variables introduce complexity around multi-threaded / multi-process Python programs and are generally considered poor software engineering practice.

We also discard any solution that has the possibility of silently incorrect behavior. For example if the user forgets to specify whether we are in the training phase. Silently incorrect behavior is a critical error that could lead to weeks of wasted researcher time. Accordingly, we discard any solution that entails a default value for the \texttt{training} argument.

We therefore decided that a required \texttt{training} keyword-only argument that must be passed recursively throughout every \texttt{SequenceLayer} API that could produce training-specific logic was the simplest way to solve the problem with zero risk of accidental misconfiguration. The cost of threading the \texttt{training} argument is a small price to pay, if ugly.

\subsection{Configuration \texttt{dataclass}es}

In this section we discuss layer configuration, and the approach taken in the JAX implementation of representing all layer configuration as nested Python dataclasses inheriting from the \texttt{SequenceLayerConfig} type. 

\begin{figure*}[h]
\begin{tcbitemize}[raster columns=1,
    raster force size=false,
    raster column 1/.style={width=1.0\textwidth},
    raster equal height,colframe=black!50,colback=white,colbacktitle=black!50,fonttitle=\rmfamily,left=1pt,right=1pt,top=1pt,bottom=1pt,
]
\tcbitem[title={Code}]
\begin{tcblisting}{
    listing engine=minted,
    minted style=friendly,
    minted language=python,
    listing only,
    left=0pt,right=0pt,top=0pt,bottom=0pt,
    boxrule=0pt,
    colback=white,
}
import dataclasses
import sequence_layers.jax as sl

class ProjectAndAdd42(sl.Stateless):
  
  @dataclasses.dataclass(frozen=True)
  class Config(sl.SequenceLayerConfig):
    num_units: int
    
    # An optional name for the layer.
    name: str | None
    
    def make(self) -> 'ProjectAndAdd42':
      return ProjectAndAdd42(self, name=self.name)
      
  config: Config
  
  def setup(self) -> None:
    self.input_projection = sl.Dense.Config(self.config.num_units).make()
    self.output_projection = sl.Dense.Config(self.config.num_units).make()

  def layer(
    self, 
    x: sl.Sequence, 
    *, 
    training: bool, 
    constants: sl.Constants | None = None
  ) -> sl.Sequence:
    y = self.input_projection.layer(x, training=training)
    y = y.apply_values(lambda v: v + 42)
    return self.output_projection.layer(x, training=training)
\end{tcblisting}

\end{tcbitemize}
  \caption{The configuration dataclass pattern in JAX SequenceLayers.}
  \label{fig:configuration-dataclasses}
\end{figure*}

In contrast, the TensorFlow implementation passes all configuration as arguments to an \texttt{\_\_init\_\_} method, as a normal Python program does. We maintain a parallel protocol buffer \citep{dean2008protobuf} library which mirrors the arguments to \texttt{\_\_init\_\_} for constructing an arbitrary composition of \texttt{SequenceLayer}s for use in other software modules.

The JAX implementation represents our evolved thinking on the topic of layer configuration, and is our recommended design pattern for implementations of the SequenceLayer design.

The key purposes that configuration serve is to provide a specification of a \texttt{SequenceLayer} to construct, without actually creating any objects that might allocate arrays (potentially creating wasted resources or requiring an accelerator to be present). 

\subsubsection{Benefits}
\begin{itemize}
    \item \textbf{Compile-time constants are externalized.} Separating the layer configuration from the layer state (submodules, trainable parameters) enables a clean separation between hyperparameters (Python values, numpy arrays) and JAX objects (Flax \texttt{nn.Module}s and \texttt{jax.Array}s). This distinction between static (compile-time constants) values and dynamic (tracers / symbolic arrays) values is useful (for example, a statically known value can be used to invoke more efficient implementations of an algorithm). We therefore think isolating static and dynamic values reduces the mental burden of having to verify whether a specific value is always statically known.
    \item \textbf{Modularity and composition at SequenceLayer call sites.} The \texttt{SequenceLayerConfig} type is essentially a thunk\footnote{\url{https://en.wikipedia.org/wiki/Thunk}} (\texttt{Callable[[], SequenceLayer]}) with the added benefit that it can be arbitrarily nested and composed within other \texttt{SequenceLayer} types, for example, multiple \texttt{SequenceLayerConfig}s can be chained via \texttt{Serial.Config}. 
    \item \textbf{Passing a description instead of an instance.} In the event that layer construction entails potentially heavyweight work, passing a description of a layer to be built can avoid accidentally wasted work or errors arising from instantiating a layer outside of its intended environment.
\end{itemize}

\subsubsection{Disadvantages}
\begin{itemize}
    \item \textbf{Increased boilerplate.} The nested configuration can increase the feeling of \texttt{SequenceLayer} implementations being bloated and heavyweight, while we are aiming for the opposite.
    \item \textbf{Copypasta mistakes.} From years of researchers authoring \texttt{SequenceLayer}s, we have found that the boilerplate entailed by the nested \texttt{Config} dataclass pattern can lead to researchers copy-pasting pre-existing layers then modifying them. A common error that the researcher can make is forgetting to update the \texttt{make} method of the config dataclass, which returns an instance of the layer the code was copied from instead of the new layer. This has happened often enough that we consider it a real and unfortunate downside of this approach despite the issue ultimately being user error.
\end{itemize}

\section{Implementation and Deployment}

\subsection{Supported Frameworks}

In our initial release, we demonstrate the SequenceLayers design in JAX (using Flax \texttt{nn.Module}s; \citealp{heek2024flax}) and TensorFlow 2. The source code is available on GitHub,\footnote{\url{https://github.com/google/sequence-layers}} and published on the Python Package Index (PyPI).\footnote{\url{https://pypi.org/project/sequence-layers/}}

See \Cref{appendix:jax-layers} for a complete list of layers provided in the JAX implementation at the time of this writing. The TensorFlow version has layers that have not yet been ported to JAX (e.g., \texttt{StyleToken}, \citealp{wang2018styletokensunsupervisedstyle}) and some that have not been backported to TensorFlow (e.g., streaming cross attention layers). This reflects the chronology of our development. SequenceLayers was initially developed as a TensorFlow library, and was ported to JAX as internal research at Google shifted towards JAX.

\subsection{Usage in Large Research Codebases}

At Google, SequenceLayers has proven a valuable tool for accelerating research velocity, enabling teams to quickly iterate on the architecture of their models without having to radically change their training and model code as architecture details change. 

SequenceLayers has been used to abstract architectural details in:
\begin{itemize}[topsep=1pt,itemsep=1pt]
    \item Classifiers
    \item Contrastive / distance metric learning models.
    \item Regression models.
    \item Probabilistic models (autoregressive models, normalizing flows, diffusion, VAEs, GANs).
\end{itemize}

across a wide variety of tasks:

\begin{itemize}[topsep=1pt,itemsep=1pt]
    \item Audio / speech classification.
    \item Image classification.
    \item Contextualized word embedding.
    \item Text-to-speech synthesis.
    \item Speech and phoneme recognition.
    \item Speech translation.
    \item Speech vocoding.
    \item Audio tokenization and synthesis.
    \item Real-time music synthesis.
    \item Video understanding.
    \item Language modeling.
\end{itemize}

The composition capabilities of SequenceLayers has enabled the creation of shared repositories of SequenceLayer definitions for varied purposes. These serve as a building block library that have enabled teams building the above models to easily share code and pre-trained models. Since layer authors adhere to the same API and test their layers for compliance with \texttt{verify\_contract}, every newly authored \texttt{SequenceLayer} is easily reusable and composable.

At time of writing, published and open-sourced work built with SequenceLayers include:

\begin{itemize}[topsep=1pt,itemsep=1pt]
    \item Gemma 3n's streaming audio encoder\footnote{\url{https://deepmind.google/models/gemma/gemma-3n}}
    \item DolphinGemma\footnote{\url{https://deepmind.google/models/gemma/dolphingemma}}
    \item \textit{Robust and Unbounded Length Generalization in Autoregressive Transformer-Based Text to Speech} (Very Attentive Tacotron; \citealp{battenberg2025vat})
    \item \textit{Source Separation by Flow Matching} \citep{scheibler2025sourceseparationflowmatching}
    \item \textit{Learning the Joint Distribution of Two Sequences using Little or No Paired Data} \citep{mariooryad2022learningjointdistributionsequences}
    \item \textit{Speaker Generation} \citep{stanton2021speakergeneration}
    \item \textit{Wave-Tacotron: Spectrogram-free End-to-End Text-to-Speech Synthesis} \citep{weiss2021wavetacotronspectrogramfreeendtoendtexttospeech}
    \item Re-implementations of Tacotron \citep{wang2017tacotron}
\end{itemize}

\section{Discussion}

\subsection{Eliminating the Research-to-Production Tax}

Scalable machine learning serving systems benefit greatly from abstraction and modularity. TensorFlow Serving \citep{olston2017tfserving} is able to serve \textit{any} TensorFlow saved model due to the dual abstractions of the \textbf{computation graph} and \textbf{signatures} that TensorFlow provides. Within Google, no additional infrastructure or model changes are needed to serve a TensorFlow saved model, since it is a well-supported abstraction.

However, TensorFlow signatures are stateless; No state is preserved across requests. The only way to deploy a streaming model on TensorFlow Serving is to plumb the state in and out of the model on every step, which is inefficient and unsuitable for large-scale deployment of models with gigabytes of state, such as Transformer KV caches in billion parameter models or low-latency / realtime scenarios where every microsecond counts.

At Google, serving streaming models at scale has typically necessitated completely custom serving systems designed for the one-off serving of a specific class of model (for example, streaming speech recognition). We refer to the creation of custom serving infrastructure or porting of models from the training implementation into a streamable implementation for serving collectively as \textbf{the research-to-production tax}.

SequenceLayers provides the API abstraction needed to eliminate the research-to-production tax for most classes of streaming models. At Google, SequenceLayers support has been integrated into standard serving infrastructure, enabling any model built with SequenceLayers to be deployed in production with no model rewriting or custom serving infrastructure.

\subsection{Dynamic Stream Processing Pipelines}

Many stream processing systems such as MediaPipe \citep{lugaresi2019mediapipeframeworkbuildingperception} enable dynamically clocked computation graphs where each node in the computation graph can consume and produce output at different times and rates. For example, nodes in a graph can buffer their inputs while producing no outputs, or block until input from multiple sinks connected to the node are ready before producing an output.

SequenceLayers is not capable of these types of dynamic computations due to the constraints of compiling programs for JAX / TensorFlow to XLA for execution on TPUs and GPUs. These constraints limit SequenceLayers' applicability to these types of problems; however, we suggest that SequenceLayers provides a complementary feature set to dynamic computation graphs, as individual nodes in a dynamic computation graph can be implemented as a SequenceLayer program running on an accelerator.

\section{Conclusion}

In this report we introduced SequenceLayers, a neural network layer API and library for sequence modeling designed from the ground up to enable seamless composition and streaming inference. We outlined its core features and how they arise from our design choices, and describe our existing implementation's layers and available architecture definitions.

This library has proved very useful to Google's research and deployment of streaming neural networks.

We look forward to your use!




\bibliography{output}
\bibliographystyle{tmlr}

\appendix 

\section{SequenceLayers Available in JAX}
\label{appendix:jax-layers}

At time of writing, we support a broad range of common layers and combinators.

\subsection{Combinator Layers}

\begin{table}[H]
\centering
\begin{tabular}{|l|p{0.7\textwidth}|}
\hline
\textbf{Name} & \textbf{Description} \\
\hline
\texttt{Bidirectional} & Processes a sequence with separate forward and backward layers and combines their outputs. \\
\hline
\texttt{Blockwise} & Processes another layer in fixed-size blocks. \\
\hline
\texttt{CheckpointGradient} & Wraps a layer with a gradient checkpoint to save memory during training. \\
\hline
\texttt{Parallel} & Applies multiple layers to the same input in parallel and combines their outputs. \\
\hline
\texttt{ParallelChannels} & Applies a single shared layer to different groups of channels in the input sequence. \\
\hline
\texttt{Repeat} & Applies a single layer multiple times sequentially in a loop. \\
\hline
\texttt{Residual} & Creates a residual connection around a sequence of layers, adding the input (with an optional shortcut layer applied) to the output. \\
\hline
\texttt{Serial} & Processes an input sequence through a series of layers, one after the other. \\
\hline
\texttt{SerialModules} & Similar to \texttt{Serial}, but for pre-constructed layer modules. \\
\hline
\end{tabular}
\caption{A summary of the available combinator layers.}
\label{table:layers-combinators}
\end{table}

\subsection{Dense and Linear Layers}

\begin{table}[H]
\centering
\begin{tabular}{|l|p{0.7\textwidth}|}
\hline
\textbf{Name} & \textbf{Description} \\
\hline
\texttt{Add} & Adds a constant value or array to the input sequence. \\
\hline
\texttt{Affine} & Applies a learnable affine transformation (scale and bias) to the input. \\
\hline
\texttt{Dense} & A standard fully-connected dense layer that operates on the final dimension of the channel shape. \\
\hline
\texttt{DenseShaped} & A dense layer that transforms the input channel shape to a specified output channel shape. \\
\hline
\texttt{EinsumDense} & A dense layer that uses an einsum equation to define the transformation between input and output shapes, for example \texttt{...ab,ac->...bc}. \\
\hline
\texttt{Embedding} & Computes learned vector embeddings for integer-coded inputs. \\
\hline
\texttt{EmbeddingTranspose} & A shared-weight transpose of an embedding layer, used for output projection. \\
\hline
\texttt{OneHot} & Computes one-hot vectors for integer-coded inputs. \\
\hline
\texttt{MaskedDense} & A causally-masked dense layer where each output timestep is a linear projection of all input timesteps at or before the current timestep. \\
\hline
\texttt{Scale} & Scales the input sequence by a constant value or array. \\
\hline
\texttt{SequenceDense} & A dense layer where a different projection is applied for each timestep. \\
\hline
\texttt{SequenceEmbedding} & An embedding layer where a different embedding table is used for each timestep. \\
\hline

\end{tabular}
\caption{A summary of the available dense and linear layers.}
\label{table:layers-dense}
\end{table}

\subsection{Attention Layers}

\begin{table}[H]
\centering
\begin{tabular}{|l|p{0.5\textwidth}|}
\hline
\textbf{Name} & \textbf{Description} \\
\hline
\texttt{DotProductSelfAttention} & A multi-headed dot-product self-attention layer with configurable causal masking. \\
\hline
\texttt{LocalDotProductSelfAttention} & Identical to \texttt{DotProductSelfAttention} in step-wise mode, but with an efficient layer-wise implementation of sliding window attention. \\
\hline
\texttt{DotProductAttention} & A standard cross-attention layer that attends to a source sequence from the \texttt{constants} dictionary (\Cref{sec:constants}). \\
\hline
\texttt{StreamingDotProductAttention} & A cross-attention layer that assumes each call to the \texttt{step} method has a different slice of the source sequence provided in the \texttt{constants} dictionary (\Cref{sec:constants}). \\
\hline
\texttt{StreamingLocalDotProductAttention} & Identical to \texttt{StreamingDotProductAttention} in step-wise mode, but with an efficient layer-wise implementation of sliding window attention. \\
\hline
\texttt{GmmAttention} & A cross-attention layer that uses a Gaussian Mixture Model to determine where to attend to the source sequence from the \texttt{constants} dictionary (\Cref{sec:constants}). Supports monotonic constraints on the location of each component of the mixture. \\
\hline
\end{tabular}
\caption{A summary of the available attention layers.}
\label{table:layers-attention}
\end{table}

The following relative position embedding schemes are supported. See \Cref{table:layers-position} for position embedding layers which can be used as an alternative to these relative embeddings.

\begin{table}[H]
\centering
\begin{tabular}{|l|p{0.5\textwidth}|}
\hline
\textbf{Name} & \textbf{Description} \\
\hline
\texttt{ShawRelativePositionEmbedding} & Computes query-dependent relative position embeddings as described by \citet{shaw2018relative}. \\
\hline
\texttt{T5RelativePositionEmbedding} & Computes relative position biases in the manner of the T5 Transformer \citep{raffel2020t5}. \\
\hline
\texttt{TransformerXLRelativePositionEmbedding} & Computes relative position embeddings in the manner of Transformer-XL \citep{dai2019transformerxl}. \\
\hline
\end{tabular}
\caption{A summary of the available relative position embedding layers.}
\label{table:layers-relative-position-embeddings}
\end{table}

\subsection{Convolution-like Layers}

\begin{table}[H]
\centering
\begin{tabular}{|l|p{0.7\textwidth}|}
\hline
\textbf{Name} & \textbf{Description} \\
\hline
\texttt{Conv1D} & A 1D strided or dilated convolution layer. \\
\hline
\texttt{Conv1DTranspose} & A 1D transpose convolution layer for upsampling. \\
\hline
\texttt{Conv2D} & A 2D strided or dilated convolution layer, with the first dimension treated as time. \\
\hline
\texttt{Conv2DTranspose} & A 2D transpose convolution layer for upsampling, with the first dimension treated as time. \\
\hline
\texttt{Conv3D} & A 3D strided or dilated convolution layer, with the first dimension treated as time. \\
\hline
\texttt{DepthwiseConv1D} & A 1D depthwise convolution layer where each input channel is convolved with its own set of filters. \\
\hline
\texttt{Downsample1D} & Downsamples the sequence along the time dimension by taking every Nth element. \\
\hline
\texttt{Upsample1D} & Upsamples the sequence along the time dimension by repetition. \\
\hline
\texttt{Upsample2D} & Upsamples the sequence along the time and one spatial dimension by repetition. \\
\hline
\end{tabular}
\caption{A summary of the available convolution layers.}
\label{table:layers-convolution}
\end{table}

\subsection{DSP Layers}

\begin{table}[H]
\centering
\begin{tabular}{|l|p{0.7\textwidth}|}
\hline
\textbf{Name} & \textbf{Description} \\
\hline
\texttt{Delay} & Delays the input sequence by a specified number of timesteps, padding with invalid timesteps. \\
\hline
\texttt{FFT} & Applies a Fast Fourier Transform (FFT) along a specified axis of the input. \\
\hline
\texttt{Frame} & Creates a sequence of overlapping frames from an input sequence. \\
\hline
\texttt{IFFT} & Applies an Inverse Fast Fourier Transform (IFFT) along a specified axis. \\
\hline
\texttt{IRFFT} & Applies an Inverse Real Fast Fourier Transform (IRFFT) to produce a real-valued output. \\
\hline
\texttt{InverseSTFT} & Computes the inverse Short-time Fourier Transform, reconstructing a signal from its spectrogram. \\
\hline
\texttt{LinearToMelSpectrogram} & Converts a linear-scale spectrogram to the mel scale. \\
\hline
\texttt{Lookahead} & Drops a specified number of initial timesteps from the input sequence. \\
\hline
\texttt{OverlapAdd} & Reconstructs a signal by overlapping and adding framed windows. \\
\hline
\texttt{RFFT} & Applies a Real Fast Fourier Transform (RFFT) for real-valued inputs. \\
\hline
\texttt{STFT} & Computes the Short-time Fourier Transform of an input signal. \\
\hline
\texttt{Window} & Applies a window function (e.g., Hann) to the input sequence along a specified axis. \\
\hline
\end{tabular}
\caption{A summary of the available DSP layers.}
\label{table:layers-dsp}
\end{table}

\subsection{Recurrent Layers}

\begin{table}[H]
\centering
\begin{tabular}{|l|p{0.7\textwidth}|}
\hline
\textbf{Name} & \textbf{Description} \\
\hline
\texttt{LSTM} & A standard Long Short-Term Memory (LSTM) layer. \\
\hline
\texttt{RGLRU} & A Real-Gated Linear Recurrent Unit (RG-LRU) layer, as used in the Griffin architecture. \\
\hline
\end{tabular}
\caption{A summary of the available recurrent layers.}
\label{table:layers-recurrent}
\end{table}

\subsection{Utility Layers}

\begin{table}[H]
\centering
\begin{tabular}{|l|p{0.7\textwidth}|}
\hline
\textbf{Name} & \textbf{Description} \\
\hline

\texttt{ApplySharding} & Applies sharding annotations to the sequence's values and mask. \\
\hline
\texttt{Argmax} & Computes the argmax along the last dimension of the input sequence. \\
\hline
\texttt{CheckpointName} & Wraps the layer with a JAX gradient checkpoint name for debugging. \\
\hline
\texttt{Dropout} & Applies dropout to the input sequence during training. \\
\hline
\texttt{Emit} & An identity layer that emits its input for debugging purposes. \\
\hline
\texttt{GradientClipping} & An identity function that clips the gradient's value during backpropagation. \\
\hline
\texttt{Lambda} & Wraps a stateless Python lambda function as a SequenceLayer. \\
\hline
\texttt{Logging} & A debugging layer that prints information about its inputs during execution. \\
\hline
\texttt{OptimizationBarrier} & Applies a JAX optimization barrier to prevent operator fusion. \\
\hline

\end{tabular}
\caption{A summary of the available utility layers.}
\label{table:layers-simple}
\end{table}

\subsection{Conditioning Layers}

\begin{table}[H]
\centering
\begin{tabular}{|l|p{0.7\textwidth}|}
\hline
\textbf{Name} & \textbf{Description} \\
\hline
\texttt{Conditioning} & Applies time-synchronized conditioning to an input sequence using a conditioning sequence in the \texttt{constants} dictionary (\Cref{sec:constants}). \\
\hline
\end{tabular}
\caption{A summary of the available conditioning layers.}
\label{table:layers-conditioning}
\end{table}

\subsection{Shape and Type Manipulation Layers}

\begin{table}[H]
\centering
\begin{tabular}{|l|p{0.7\textwidth}|}
\hline
\textbf{Name} & \textbf{Description} \\
\hline
\texttt{Cast} & Casts the input sequence to a specified data type. \\
\hline
\texttt{EinopsRearrange} & Rearranges the channel dimensions of the input using an \texttt{einops.rearrange} equation. \\
\hline
\texttt{ExpandDims} & Adds new dimensions of size 1 to the channel shape. \\
\hline
\texttt{Flatten} & Flattens all channel dimensions into a single dimension. \\
\hline
\texttt{GlobalEinopsRearrange} & Rearranges both the time and channel dimensions using an \texttt{einops.rearrange} equation. \\
\hline
\texttt{GlobalReshape} & Reshapes both the time and channel dimensions of the sequence. \\
\hline
\texttt{MoveAxis} & Moves channel axes to new positions. \\
\hline
\texttt{Reshape} & Reshapes the channel dimensions of the input sequence. \\
\hline
\texttt{Slice} & Slices the channel dimensions of the input sequence. \\
\hline
\texttt{Squeeze} & Removes channel dimensions of size 1. \\
\hline
\texttt{SwapAxes} & Swaps two channel axes of the input. \\
\hline
\texttt{Transpose} & Permutes the channel dimensions of the input. \\
\hline
\end{tabular}
\caption{A summary of the available shape and type manipulation layers.}
\label{table:layers-shape-type-manipulation}
\end{table}

\subsection{Activation and Pointwise Layers}

\begin{table}[H]
\centering
\begin{tabular}{|l|p{0.7\textwidth}|}
\hline
\textbf{Name} & \textbf{Description} \\
\hline

\texttt{Abs} & Takes the element-wise absolute value of the input sequence. \\
\hline
\texttt{Elu} & Applies the Exponential Linear Unit (ELU) activation function. \\
\hline
\texttt{Exp} & Applies the element-wise exponential function to the input. \\
\hline
\texttt{GatedLinearUnit} & A Gated Linear Unit (GLU) that halves the channel dimension. \\
\hline
\texttt{GatedTanhUnit} & A Gated Tanh Unit that halves the channel dimension. \\
\hline
\texttt{GatedUnit} & A generalized gated unit that combines two halves of the input with activations. \\
\hline
\texttt{Gelu} & Applies the Gaussian Error Linear Unit (GELU) activation function. \\
\hline
\texttt{Identity} & An identity layer that passes its input through unchanged. \\
\hline
\texttt{LeakyRelu} & Applies the Leaky Rectified Linear Unit (Leaky ReLU) activation function. \\
\hline
\texttt{Log} & Applies the element-wise natural logarithm to the input. \\
\hline
\texttt{MaskInvalid} & Replaces invalid (masked) timesteps in the sequence with zeros. \\
\hline
\texttt{Maximum} & Performs an element-wise clip with a specified maximum value. \\
\hline
\texttt{Minimum} & Performs an element-wise clip with a specified minimum value. \\
\hline
\texttt{Mod} & Computes the element-wise remainder of division by a specified divisor. \\
\hline
\texttt{PRelu} & Applies a Parametric ReLU where the negative slope is a learnable parameter. \\
\hline
\texttt{Power} & Raises the input sequence to a specified power. \\
\hline
\texttt{Relu} & Applies the Rectified Linear Unit (ReLU) activation function. \\
\hline
\texttt{Sigmoid} & Applies the sigmoid activation function. \\
\hline
\texttt{Softmax} & Applies the softmax function along a specified channel axis. \\
\hline
\texttt{Softplus} & Applies the softplus activation function. \\
\hline
\texttt{Swish} & Applies the Swish activation function. \\
\hline
\texttt{Tanh} & Applies the hyperbolic tangent activation function. \\
\hline
\end{tabular}
\caption{A summary of the available activation and pointwise layers.}
\label{table:layers-pointwise}
\end{table}

\subsection{Normalization Layers}

\begin{table}[H]
\centering
\begin{tabular}{|l|p{0.7\textwidth}|}
\hline
\textbf{Name} & \textbf{Description} \\
\hline
\texttt{BatchNormalization} & Applies batch normalization, normalizing across the batch and time dimensions for each feature. \\
\hline
\texttt{GroupNormalization} & Applies group normalization, dividing channels into groups and normalizing within each group. \\
\hline
\texttt{LayerNormalization} & Applies layer normalization, normalizing over the feature axes for each item in the batch and time. \\
\hline
\texttt{RMSNormalization} & Applies Root Mean Square (RMS) normalization, a simplified version of layer normalization without mean-centering. \\
\hline
\end{tabular}
\caption{A summary of the available normalization layers.}
\label{table:layers-normalization}
\end{table}

\subsection{Pooling Layers}

\begin{table}[H]
\centering
\begin{tabular}{|l|p{0.7\textwidth}|}
\hline
\textbf{Name} & \textbf{Description} \\
\hline
\texttt{AveragePooling1D} & A 1D pooling layer that reduces temporal resolution by taking the average over a sliding window. \\
\hline
\texttt{AveragePooling2D} & A 2D pooling layer that reduces temporal and spatial resolution by taking the average over a sliding window. \\
\hline
\texttt{MaxPooling1D} & A 1D pooling layer that reduces temporal resolution by taking the maximum over a sliding window. \\
\hline
\texttt{MaxPooling2D} & A 2D pooling layer that reduces temporal and spatial resolution by taking the maximum over a sliding window. \\
\hline
\texttt{MinPooling1D} & A 1D pooling layer that reduces temporal resolution by taking the minimum over a sliding window. \\
\hline
\texttt{MinPooling2D} & A 2D pooling layer that reduces temporal and spatial resolution by taking the minimum over a sliding window. \\
\hline
\end{tabular}
\caption{A summary of the available pooling layers.}
\label{table:layers-pooling}
\end{table}

\subsection{Position Embedding Layers}

\begin{table}[H]
\centering
\begin{tabular}{|l|p{0.5\textwidth}|}
\hline
\textbf{Name} & \textbf{Description} \\
\hline
\texttt{AddTimingSignal} & Adds sinusoidal timing signals of varying frequencies to the input channels. \\
\hline
\texttt{ApplyRotaryPositionalEncoding} & Applies Rotary Positional Encodings (RoPE) to the sequence to provide relative position information. \\
\hline
\end{tabular}
\caption{A summary of the available position embedding layers.}
\label{table:layers-position}
\end{table}

\end{document}

%% file: math_commands.tex
\usepackage{amsmath,amsfonts,bm}

\def\eqref#1{equation~\ref{#1}}

\def\1{\bm{1}}

\DeclareMathAlphabet{\mathsfit}{\encodingdefault}{\sfdefault}{m}{sl}
\SetMathAlphabet{\mathsfit}{bold}{\encodingdefault}{\sfdefault}{bx}{n}













%% file: output.bbl
\begin{thebibliography}{28}
\providecommand{\natexlab}[1]{#1}
\providecommand{\url}[1]{\texttt{#1}}
\expandafter\ifx\csname urlstyle\endcsname\relax
  \providecommand{\doi}[1]{doi: #1}\else
  \providecommand{\doi}{doi: \begingroup \urlstyle{rm}\Url}\fi

\bibitem[Abadi et~al.(2016)Abadi, Barham, Chen, Chen, Davis, Dean, Devin,
  Ghemawat, Irving, Isard, Kudlur, Levenberg, Monga, Moore, Murray, Steiner,
  Tucker, Vasudevan, Warden, Wicke, Yu, and Zheng]{abadi2016tensorflow}
Mart{\'{\i}}n Abadi, Paul Barham, Jianmin Chen, Zhifeng Chen, Andy Davis,
  Jeffrey Dean, Matthieu Devin, Sanjay Ghemawat, Geoffrey Irving, Michael
  Isard, Manjunath Kudlur, Josh Levenberg, Rajat Monga, Sherry Moore,
  Derek~Gordon Murray, Benoit Steiner, Paul~A. Tucker, Vijay Vasudevan, Pete
  Warden, Martin Wicke, Yuan Yu, and Xiaoqiang Zheng.
\newblock Tensor{F}low: {A} system for large-scale machine learning.
\newblock In \emph{{OSDI}}, pp.\  265--283. {USENIX} Association, 2016.

\bibitem[Bahdanau et~al.(2015)Bahdanau, Cho, and
  Bengio]{bahdanau2016neuralmachinetranslationjointly}
Dzmitry Bahdanau, Kyunghyun Cho, and Yoshua Bengio.
\newblock Neural machine translation by jointly learning to align and
  translate.
\newblock In \emph{{ICLR}}, 2015.

\bibitem[Battenberg et~al.(2025)Battenberg, Skerry{-}Ryan, Stanton, Mariooryad,
  Shannon, Salazar, and Kao]{battenberg2025vat}
Eric Battenberg, R.~J. Skerry{-}Ryan, Daisy Stanton, Soroosh Mariooryad, Matt
  Shannon, Julian Salazar, and David Kao.
\newblock Robust and unbounded length generalization in autoregressive
  transformer-based text-to-speech.
\newblock In \emph{{NAACL} (Long Papers)}, pp.\  11789--11806. Association for
  Computational Linguistics, 2025.

\bibitem[Chollet et~al.(2018)]{chollet2018keras}
Fran{\c{c}}ois Chollet et~al.
\newblock Keras: The {P}ython deep learning library.
\newblock \emph{Astrophysics source code library}, pp.\  ascl--1806, 2018.

\bibitem[Dai et~al.(2019)Dai, Yang, Yang, Carbonell, Le, and
  Salakhutdinov]{dai2019transformerxl}
Zihang Dai, Zhilin Yang, Yiming Yang, Jaime~G. Carbonell, Quoc~Viet Le, and
  Ruslan Salakhutdinov.
\newblock Transformer-{XL}: Attentive language models beyond a fixed-length
  context.
\newblock In \emph{{ACL} {(1)}}, pp.\  2978--2988. Association for
  Computational Linguistics, 2019.

\bibitem[Dean et~al.(2008)Dean, Ghemawat, et~al.]{dean2008protobuf}
Jeff Dean, Sanjay Ghemawat, et~al.
\newblock Protocol buffers, 2008.

\bibitem[Frostig et~al.(2018)Frostig, Johnson, and Leary]{frostig2018jax}
Roy Frostig, Matthew~James Johnson, and Chris Leary.
\newblock Compiling machine learning programs via high-level tracing.
\newblock \emph{Systems for Machine Learning}, 4\penalty0 (9), 2018.

\bibitem[Hannun et~al.(2023)Hannun, Digani, Katharopoulos, and
  Collobert]{mlx2023}
Awni Hannun, Jagrit Digani, Angelos Katharopoulos, and Ronan Collobert.
\newblock {MLX}: Efficient and flexible machine learning on {A}pple silicon,
  2023.
\newblock URL \url{https://github.com/ml-explore}.

\bibitem[He et~al.(2016)He, Zhang, Ren, and Sun]{he2015resnet}
Kaiming He, Xiangyu Zhang, Shaoqing Ren, and Jian Sun.
\newblock Deep residual learning for image recognition.
\newblock In \emph{{CVPR}}, pp.\  770--778. {IEEE} Computer Society, 2016.

\bibitem[Heek et~al.(2024)Heek, Levskaya, Oliver, Ritter, Rondepierre, Steiner,
  and van {Z}ee]{heek2024flax}
Jonathan Heek, Anselm Levskaya, Avital Oliver, Marvin Ritter, Bertrand
  Rondepierre, Andreas Steiner, and Marc van {Z}ee.
\newblock {F}lax: A neural network library and ecosystem for {JAX}, 2024.
\newblock URL \url{http://github.com/google/flax}.

\bibitem[Lugaresi et~al.(2019)Lugaresi, Tang, Nash, McClanahan, Uboweja, Hays,
  Zhang, Chang, Yong, Lee, Chang, Hua, Georg, and
  Grundmann]{lugaresi2019mediapipeframeworkbuildingperception}
Camillo Lugaresi, Jiuqiang Tang, Hadon Nash, Chris McClanahan, Esha Uboweja,
  Michael Hays, Fan Zhang, Chuo{-}Ling Chang, Ming~Guang Yong, Juhyun Lee,
  Wan{-}Teh Chang, Wei Hua, Manfred Georg, and Matthias Grundmann.
\newblock Mediapipe: {A} framework for building perception pipelines.
\newblock \emph{CoRR}, abs/1906.08172, 2019.

\bibitem[Mariooryad et~al.(2022)Mariooryad, Shannon, Ma, Bagby, Kao, Stanton,
  Battenberg, and
  Skerry{-}Ryan]{mariooryad2022learningjointdistributionsequences}
Soroosh Mariooryad, Matt Shannon, Siyuan Ma, Tom Bagby, David Kao, Daisy
  Stanton, Eric Battenberg, and R.~J. Skerry{-}Ryan.
\newblock Learning the joint distribution of two sequences using little or no
  paired data.
\newblock \emph{CoRR}, abs/2212.03232, 2022.

\bibitem[Mohiuddin et~al.(2019)]{trax2019}
Afroz Mohiuddin et~al.
\newblock Trax — deep learning with clear code and speed, 2019.
\newblock URL \url{https://github.com/google/trax}.

\bibitem[Olston et~al.(2017)Olston, Li, Harmsen, Soyke, Gorovoy, Lao, Fiedel,
  Ramesh, and Rajashekhar]{olston2017tfserving}
Christopher Olston, Fangwei Li, Jeremiah Harmsen, Jordan Soyke, Kiril Gorovoy,
  Li~Lao, Noah Fiedel, Sukriti Ramesh, and Vinu Rajashekhar.
\newblock Tensor{F}low-{S}erving: Flexible, high-performance {ML} serving.
\newblock In \emph{Workshop on ML Systems at NIPS 2017}, 2017.

\bibitem[Paszke et~al.(2019)Paszke, Gross, Massa, Lerer, Bradbury, Chanan,
  Killeen, Lin, Gimelshein, Antiga, Desmaison, K{\"{o}}pf, Yang, DeVito,
  Raison, Tejani, Chilamkurthy, Steiner, Fang, Bai, and
  Chintala]{paszke2019pytorch}
Adam Paszke, Sam Gross, Francisco Massa, Adam Lerer, James Bradbury, Gregory
  Chanan, Trevor Killeen, Zeming Lin, Natalia Gimelshein, Luca Antiga, Alban
  Desmaison, Andreas K{\"{o}}pf, Edward~Z. Yang, Zachary DeVito, Martin Raison,
  Alykhan Tejani, Sasank Chilamkurthy, Benoit Steiner, Lu~Fang, Junjie Bai, and
  Soumith Chintala.
\newblock Py{T}orch: An imperative style, high-performance deep learning
  library.
\newblock In \emph{NeurIPS}, pp.\  8024--8035, 2019.

\bibitem[Raffel et~al.(2020)Raffel, Shazeer, Roberts, Lee, Narang, Matena,
  Zhou, Li, and Liu]{raffel2020t5}
Colin Raffel, Noam Shazeer, Adam Roberts, Katherine Lee, Sharan Narang, Michael
  Matena, Yanqi Zhou, Wei Li, and Peter~J. Liu.
\newblock Exploring the limits of transfer learning with a unified text-to-text
  transformer.
\newblock \emph{J. Mach. Learn. Res.}, 21:\penalty0 140:1--140:67, 2020.

\bibitem[Roberts et~al.(2023)Roberts, Chung, Mishra, Levskaya, Bradbury, Andor,
  Narang, Lester, Gaffney, Mohiuddin, Hawthorne, Lewkowycz, Salcianu, van Zee,
  Austin, Goodman, Soares, Hu, Tsvyashchenko, Chowdhery, Bastings, Bulian,
  Garcia, Ni, Chen, Kenealy, Han, Casbon, Clark, Lee, Garrette, Lee{-}Thorp,
  Raffel, Shazeer, Ritter, Bosma, Passos, Maitin{-}Shepard, Fiedel, Omernick,
  Saeta, Sepassi, Spiridonov, Newlan, and Gesmundo]{roberts2023t5x}
Adam Roberts, Hyung~Won Chung, Gaurav Mishra, Anselm Levskaya, James Bradbury,
  Daniel Andor, Sharan Narang, Brian Lester, Colin Gaffney, Afroz Mohiuddin,
  Curtis Hawthorne, Aitor Lewkowycz, Alex Salcianu, Marc van Zee, Jacob Austin,
  Sebastian Goodman, Livio~Baldini Soares, Haitang Hu, Sasha Tsvyashchenko,
  Aakanksha Chowdhery, Jasmijn Bastings, Jannis Bulian, Xavier Garcia, Jianmo
  Ni, Andrew Chen, Kathleen Kenealy, Kehang Han, Michelle Casbon, Jonathan~H.
  Clark, Stephan Lee, Dan Garrette, James Lee{-}Thorp, Colin Raffel, Noam
  Shazeer, Marvin Ritter, Maarten Bosma, Alexandre Passos, Jeremy
  Maitin{-}Shepard, Noah Fiedel, Mark Omernick, Brennan Saeta, Ryan Sepassi,
  Alexander Spiridonov, Joshua Newlan, and Andrea Gesmundo.
\newblock Scaling up models and data with t5x and seqio.
\newblock \emph{J. Mach. Learn. Res.}, 24:\penalty0 377:1--377:8, 2023.

\bibitem[Sabne(2020)]{sabne2020xla}
Amit Sabne.
\newblock {XLA}: Compiling machine learning for peak performance, 2020.

\bibitem[Scheibler et~al.(2025)Scheibler, Hershey, Doucet, and
  Li]{scheibler2025sourceseparationflowmatching}
Robin Scheibler, John~R. Hershey, Arnaud Doucet, and Henry Li.
\newblock Source separation by flow matching.
\newblock \emph{CoRR}, abs/2505.16119, 2025.

\bibitem[Shaw et~al.(2018)Shaw, Uszkoreit, and Vaswani]{shaw2018relative}
Peter Shaw, Jakob Uszkoreit, and Ashish Vaswani.
\newblock Self-attention with relative position representations.
\newblock In \emph{{NAACL-HLT} {(2)}}, pp.\  464--468. Association for
  Computational Linguistics, 2018.

\bibitem[Shen et~al.(2019)Shen, Nguyen, Wu, Chen, Chen, Jia, Kannan, Sainath,
  Cao, Chiu, He, Chorowski, Hinsu, Laurenzo, Qin, Firat, Macherey, Gupta,
  Bapna, Zhang, Pang, Weiss, Prabhavalkar, Liang, Jacob, Liang, Lee, Chelba,
  Jean, Li, Johnson, Anil, Tibrewal, Liu, Eriguchi, Jaitly, Ari, Cherry,
  Haghani, Good, Cheng, Alvarez, Caswell, Hsu, Yang, Wang, Gonina, Tomanek,
  Vanik, Wu, Jones, Schuster, Huang, Chen, Irie, Foster, Richardson, Macherey,
  Bruguier, Zen, Raffel, Kumar, Rao, Rybach, Murray, Peddinti, Krikun,
  Bacchiani, Jablin, Suderman, Williams, Lee, Bhatia, Carlson, Yavuz, Zhang,
  McGraw, Galkin, Ge, Pundak, Whipkey, Wang, Alon, Lepikhin, Tian, Sabour,
  Chan, Toshniwal, Liao, Nirschl, and
  Rondon]{shen2019lingvomodularscalableframework}
Jonathan Shen, Patrick Nguyen, Yonghui Wu, Zhifeng Chen, Mia~Xu Chen, Ye~Jia,
  Anjuli Kannan, Tara~N. Sainath, Yuan Cao, Chung{-}Cheng Chiu, Yanzhang He,
  Jan Chorowski, Smit Hinsu, Stella Laurenzo, James Qin, Orhan Firat, Wolfgang
  Macherey, Suyog Gupta, Ankur Bapna, Shuyuan Zhang, Ruoming Pang, Ron~J.
  Weiss, Rohit Prabhavalkar, Qiao Liang, Benoit Jacob, Bowen Liang, HyoukJoong
  Lee, Ciprian Chelba, S{\'{e}}bastien Jean, Bo~Li, Melvin Johnson, Rohan Anil,
  Rajat Tibrewal, Xiaobing Liu, Akiko Eriguchi, Navdeep Jaitly, Naveen Ari,
  Colin Cherry, Parisa Haghani, Otavio Good, Youlong Cheng, Raziel Alvarez,
  Isaac Caswell, Wei{-}Ning Hsu, Zongheng Yang, Kuan{-}Chieh Wang, Ekaterina
  Gonina, Katrin Tomanek, Ben Vanik, Zelin Wu, Llion Jones, Mike Schuster,
  Yanping Huang, Dehao Chen, Kazuki Irie, George~F. Foster, John Richardson,
  Klaus Macherey, Antoine Bruguier, Heiga Zen, Colin Raffel, Shankar Kumar,
  Kanishka Rao, David Rybach, Matthew Murray, Vijayaditya Peddinti, Maxim
  Krikun, Michiel Bacchiani, Thomas~B. Jablin, Robert Suderman, Ian Williams,
  Benjamin Lee, Deepti Bhatia, Justin Carlson, Semih Yavuz, Yu~Zhang, Ian
  McGraw, Max Galkin, Qi~Ge, Golan Pundak, Chad Whipkey, Todd Wang, Uri Alon,
  Dmitry Lepikhin, Ye~Tian, Sara Sabour, William Chan, Shubham Toshniwal,
  Baohua Liao, Michael Nirschl, and Pat Rondon.
\newblock Lingvo: a modular and scalable framework for sequence-to-sequence
  modeling.
\newblock \emph{CoRR}, abs/1902.08295, 2019.

\bibitem[Stanton et~al.(2022)Stanton, Shannon, Mariooryad, Skerry{-}Ryan,
  Battenberg, Bagby, and Kao]{stanton2021speakergeneration}
Daisy Stanton, Matt Shannon, Soroosh Mariooryad, R.~J. Skerry{-}Ryan, Eric
  Battenberg, Tom Bagby, and David Kao.
\newblock Speaker generation.
\newblock In \emph{{ICASSP}}, pp.\  7897--7901. {IEEE}, 2022.

\bibitem[Sun et~al.(2024)Sun, Dong, Zhu, Huang, Wang, Ma, Zhang, Wang, and
  Wei]{sun2024yoco}
Yutao Sun, Li~Dong, Yi~Zhu, Shaohan Huang, Wenhui Wang, Shuming Ma, Quanlu
  Zhang, Jianyong Wang, and Furu Wei.
\newblock You only cache once: Decoder-decoder architectures for language
  models.
\newblock In \emph{NeurIPS}, 2024.

\bibitem[Vaswani et~al.(2017)Vaswani, Shazeer, Parmar, Uszkoreit, Jones, Gomez,
  Kaiser, and Polosukhin]{vaswani2017transformer}
Ashish Vaswani, Noam Shazeer, Niki Parmar, Jakob Uszkoreit, Llion Jones,
  Aidan~N. Gomez, Lukasz Kaiser, and Illia Polosukhin.
\newblock Attention is all you need.
\newblock In \emph{{NIPS}}, pp.\  5998--6008, 2017.

\bibitem[Wang et~al.(2017)Wang, Skerry{-}Ryan, Stanton, Wu, Weiss, Jaitly,
  Yang, Xiao, Chen, Bengio, Le, Agiomyrgiannakis, Clark, and
  Saurous]{wang2017tacotron}
Yuxuan Wang, R.~J. Skerry{-}Ryan, Daisy Stanton, Yonghui Wu, Ron~J. Weiss,
  Navdeep Jaitly, Zongheng Yang, Ying Xiao, Zhifeng Chen, Samy Bengio, Quoc~V.
  Le, Yannis Agiomyrgiannakis, Rob Clark, and Rif~A. Saurous.
\newblock Tacotron: Towards end-to-end speech synthesis.
\newblock In \emph{{INTERSPEECH}}, pp.\  4006--4010. {ISCA}, 2017.

\bibitem[Wang et~al.(2018)Wang, Stanton, Zhang, Skerry{-}Ryan, Battenberg,
  Shor, Xiao, Jia, Ren, and Saurous]{wang2018styletokensunsupervisedstyle}
Yuxuan Wang, Daisy Stanton, Yu~Zhang, R.~J. Skerry{-}Ryan, Eric Battenberg,
  Joel Shor, Ying Xiao, Ye~Jia, Fei Ren, and Rif~A. Saurous.
\newblock Style tokens: Unsupervised style modeling, control and transfer in
  end-to-end speech synthesis.
\newblock In \emph{{ICML}}, volume~80 of \emph{Proceedings of Machine Learning
  Research}, pp.\  5167--5176. {PMLR}, 2018.

\bibitem[Weiss et~al.(2021)Weiss, Skerry{-}Ryan, Battenberg, Mariooryad, and
  Kingma]{weiss2021wavetacotronspectrogramfreeendtoendtexttospeech}
Ron~J. Weiss, R.~J. Skerry{-}Ryan, Eric Battenberg, Soroosh Mariooryad, and
  Diederik~P. Kingma.
\newblock Wave-{T}acotron: Spectrogram-free end-to-end text-to-speech
  synthesis.
\newblock In \emph{{ICASSP}}, pp.\  5679--5683. {IEEE}, 2021.

\bibitem[Wolf et~al.(2019)Wolf, Debut, Sanh, Chaumond, Delangue, Moi, Cistac,
  Rault, Louf, Funtowicz, and Brew]{wolf2019huggingface}
Thomas Wolf, Lysandre Debut, Victor Sanh, Julien Chaumond, Clement Delangue,
  Anthony Moi, Pierric Cistac, Tim Rault, R{\'{e}}mi Louf, Morgan Funtowicz,
  and Jamie Brew.
\newblock Hugging{F}ace's {T}ransformers: State-of-the-art natural language
  processing.
\newblock \emph{CoRR}, abs/1910.03771, 2019.

\end{thebibliography}
